%% file: main.tex
\documentclass[lettersize,journal]{IEEEtran}
\IEEEoverridecommandlockouts
\usepackage{amsfonts}
\usepackage{bbm}
\usepackage{graphicx}
\usepackage{amsthm} 

\usepackage{xcolor}
\usepackage{float}   
\usepackage[normalem]{ulem}
\usepackage{dsfont}
\usepackage{textcomp}
\usepackage{stfloats}
\usepackage{url}
\usepackage{verbatim}
\usepackage{algorithmic}
\usepackage{algorithm}

\usepackage{array}
\usepackage{booktabs}
\usepackage{pifont}
\usepackage{comment}
\usepackage{algorithm}
\usepackage{algorithmic}

\usepackage{algorithm}

\usepackage[acronym]{glossaries}
\input{acronyms}
\usepackage{amsmath,amsfonts, amssymb}
\usepackage{bm}
\usepackage{graphicx}

\usepackage[font=small]{caption}
\usepackage[font=footnotesize]{subcaption}
\usepackage{tikz}
\usetikzlibrary{plotmarks,patterns,decorations.pathreplacing,backgrounds,calc,arrows,arrows.meta,spy,matrix}
\usepackage{tikzscale}

\usepackage{pgfplots}
\usepgfplotslibrary{colorbrewer, groupplots, patchplots}
\pgfplotsset{
    compat=newest,
    legend style={font=\footnotesize, fill opacity=0.7,  draw opacity=1, text opacity=1, draw=white!15!black, legend cell align=left, align=left}, 
    width=0.8\columnwidth, 
    scale only axis,
    height=4cm,
    yminorticks=false,
    xminorticks=false,
    label style={font=\small},
    title style={font=\small},
    tick align=outside,
    tick pos=left,
    tick style={color=black},
    tick label style={font=\footnotesize},
    grid style={line width=.1pt, draw=gray!20},
    major grid style={line width=.1pt,draw=gray!20},
    plot coordinates/math parser=false 
}
\newlength\figureheight
\newlength\figurewidth

\usepackage{multirow}
\usepackage{tablefootnote}
\usepackage{booktabs}
\usepackage{tabularx}

\def\BibTeX{{\rm B\kern-.05em{\sc i\kern-.025em b}\kern-.08em T\kern-.1667em\lower.7ex\hbox{E}\kern-.125emX}}

\newcommand{\probP}{\text{I\kern-0.15em P}}

\definecolor{darkgreen}{RGB}{0,100,0}

\begin{document}

\title{Link-Aware Energy-Frugal Continual Learning for Fault Detection in IoT Networks}
\author{Henrik C. M. Frederiksen, Junya Shiraishi,~\IEEEmembership{Member,~IEEE}, \v Cedomir~Stefanovi\' c,~\IEEEmembership{Senior Member,~IEEE},\\ Hei Victor Cheng,~\IEEEmembership{Member,~IEEE}, and Shashi Raj Pandey,~\IEEEmembership{Member,~IEEE}\\
\thanks{
The work of H. C. M. Frederiksen and S. R. Pandey was supported in parts by DFF-Forskningsprojekt1 ``NETML" with grant No. 4286-00278B. The work of J. Shiraishi was supported by European Union's Horizon Europe research and innovation funding programme under Marie Sk{\l}odowska-Curie Action (MSCA) Postdoctoral Fellowship, ``NEUTRINAI" with grant agreement No.~101151067. (Corresponding author: Henrik C. M. Frederiksen)}
\thanks{
H. C. M. Frederiksen, J. Shiraishi, C. Stefanovi\' c, and S. R. Pandey are with Department of Electronic Systems, Aalborg University, 9220 Aalborg, Denmark (Email: \{hcmf, jush, cs, srp\}@es.aau.dk). H. V. Cheng is with Department of Electrical and Computer Engineering, Aarhus University, Denmark (e-mail: hvc@ece.au.dk).
}
}

\maketitle

\begin{abstract}
The use of lightweight machine learning (ML) models in internet of things (IoT) networks enables resource-constrained IoT devices to perform on-device inference for several critical applications. However, the inference accuracy deteriorates due to the non-stationarity in the IoT environment and limited initial training data. To counteract this, the deployed models can be updated occasionally with new observed data samples. However, this approach consumes additional energy, which is undesirable for energy constrained IoT devices. This letter introduces an event-driven communication framework that strategically integrates continual learning (CL) in IoT networks for energy-efficient fault detection. Our framework enables the IoT device and the edge server (ES) to collaboratively update the lightweight ML model by adapting to the wireless link conditions for communication and the available energy budget. Evaluation on real-world datasets show that the proposed approach can outperform both periodic sampling and non-adaptive CL in terms of inference recall; our proposed approach achieves up to a 42.8\% improvement, even under tight energy and bandwidth constraint.
\end{abstract}

\section{Introduction}
\IEEEPARstart{L}{ightweight} \gls{ml} models play a crucial role in realizing a variety of future 6G \gls{iot} applications, including correct response to semantic queries~\cite{tinyairnet}, and accurate on-device inference for real-time tasks such as fault detection in an industrial setting~\cite{edgefault}.  
For example, an on-device lightweight \gls{ml} model can act as a fault detector by identifying potential faults from its own sensory inputs. This enables the \gls{iot} device to interpret complex patterns in observed sensor measurements and assess whether a potential fault event is happening in real time.  
However, lack of sufficient training data at the initial deployment phase in practical applications hinders the universal deployment of on-device inference models. Using only pre-trained supervised detectors obviously struggles with previously unseen faults, which is highly likely to suffer from misclassifications~\cite{Michau-2019}.

The \gls{cl} framework~\cite{hawk} is an attractive approach to improve the accuracy of the deployed \gls{ml} model through its continuous adaptation towards changing data distributions. To alleviate the computational burden of frequently updating the model, and issues of learning from limited faulty local data on constrained \gls{iot} devices, an \gls{es} can help iteratively update the \gls{ml} model based on the data collected from the distributed edge device.
However, it comes with two challenges: \emph{first}, collecting data and updating the model centrally adds energy costs for data transmission at the expense of delayed inference; \textit{second,} the periodicity of model updates and how it is shared to the \gls{iot} devices should adapt to changing link statistics. It is yet unclear in the existing literature, how to maintain accurate \gls{fd} while maintaining total energy consumption below the available energy budget at the \gls{iot} device, and how to adapt model compression per changing link conditions for better on-device inference accuracy. This entails designing the communication protocol akin to model and data exchange strategies, by taking into account the inherent trade-off between the total energy consumption and inference accuracy.

To address this problem, this paper introduces \gls{acord}, an event-driven energy frugal communication framework integrating \gls{cl}.
In \gls{acord} the \gls{iot} devices sends data that could contribute to the improvement of the model accuracy, only when it observes potential rare events. The \gls{es} then improves the model by applying the \gls{cl} method~\cite{hawk} and compresses the resulting model before transmitting it to the \gls{iot} devices. The framework leverages an adaptive compression method for exchanging potential fault data and updated \gls{ml} models by leveraging the estimated link condition and available energy budget.

The contributions can be summarized as follows: \emph{i}) We propose a~\gls{cl} framework for fault detection in IoT networks, designed to operate under constraints on computation, bandwidth, and energy availability; \emph{ii}) We propose a link-aware adaptation algorithm, in which we iteratively optimize data/\gls{ml} model transmission parameters for the \gls{iot} device and for the \gls{es} in order to offer high inference accuracy at the \gls{iot} device under the constraint of energy consumption; \emph{iii}) We characterize the system level performance of the proposed framework in terms of total energy consumption and inference accuracy for a variety of parameters and via an experimental study in a testbed, validate the frameworks' feasibility; \emph{iv}) We elicit the gain brought by the proposed approach compared with the two intuitive baseline schemes: Periodic Sampling and Hawk~\cite{hawk} in terms of inference accuracy under the energy constraint.

\section{System Model}\label{sec:sys_model}
We consider an online \gls{iot} \gls{fd} scenario, in which a single
battery-powered \gls{iot} device periodically monitors sensor data for detecting potential system faults and communicates with an \gls{es} via an \gls{ap}. Fault events at the \gls{iot} device are defined as an event that deviates from the desired statistical behavior~\cite{faultdefinition}, recorded when the system owner reports the system as not operating properly.

The \gls{iot} device comprises sensors, a \gls{mcu}, a wireless communication module, and a battery. The \gls{iot} device runs a lightweight \gls{fd} model for continuously monitoring the local environment/system. 
We assume a \gls{cl} setup for \gls{iot} networks, in which the deployed \gls{fd} is continuously updated with new data.  
This scenario includes a variety of practical considerations in the context of \gls{ml} model deployment, e.g., the initial model is trained only using normal data due to a lack of fault data~\cite{Michau-2019}.

The training of the \gls{ml} model is carried out in the \gls{es} by collecting the training data from the \gls{iot} device. As local training is infeasible at the \gls{iot} device due to memory and compute constraints, it needs to rely on the \gls{es} to improve the model as well as labeling for newly observed data. 
After finishing training at the \gls{es}, the improved model is shared with the \gls{iot} device through downlink communication.

 \subsection{Communication System}\label{sec:com-system}
 
In order to realize the above operation, we introduce an event-driven communication framework applying \gls{cl}. 
Time is divided into fault detection rounds. The $i$-th round is defined as the duration in which the \gls{iot} device uses the $i$-th generation of fault detection model, denoted as $\mathcal{M}^{i}$. Each round consists of four phases, as follows: 

\subsubsection{\gls{fd} phase}\label{sec:frame_phase_1}
After receiving a new \gls{ml} model, $\mathcal{M}^i$, the \gls{iot} device resumes fault monitoring using $\mathcal{M}^i$, starting a new round.
Let $\bm{x}_k^i = [x_{k, 1}^i, x_{k, 2}^i, \ldots, x_{k, N}^i]^\top \in \mathbb{R}^N$ denote the $k$-th $N$-dimensional timeseries sample observed by the \gls{iot} device in \gls{fd} round $i$. Further, let $s_k^i \in \{0,1\}$ be the corresponding ground truth label, where $s_k^i = 0$ indicates a current observation is from a device in a normal state while $s_k^i = 1$ means that the current observation is from a device in a faulty state.
After each sampling period, the model takes $\bm{x}_k^i$ as input, and based on a classification decision threshold $\tau_{\mathrm{th}}$, outputs an estimated label $\hat{s}_k^i$ as $\mathcal{M}^i(\bm{x}_k^i; \tau_{\mathrm{th}}): \bm{x}_k^i \rightarrow \hat{s}_k^i \in \{0, 1\}$, where
$\tau_{\mathrm{th}}$ controls sensitivity of $\mathcal{M}^i$; higher (lower) $\tau_{\mathrm{th}}$ requiring more (less) certainty of fault state from the model resulting in less (more) $\hat{s}_k^i = 1$ and smaller (larger) energy consumption.

A temporal correlation is assumed among subsequent timeseries samples. The time interval during which the observations are considered as correlated is referred to as coherence time of observations. 
If $\hat{s}_k^i = 1$, the \gls{iot} device accumulates a subset of data, related to the current observation with the parameter $W$, which we refer to as the context window, as $\mathcal{Q}^i \leftarrow \mathcal{Q}^i \bigcup\{\bm{x}_j^i\}_{j=k -W}^{k+W}$, i.e., along with $2W + 1$ samples. Here, the higher (lower) $W$ generally results in better (worse) predictions $\hat{s}_k^i$ for future episodes because of the increasing available training data at the \gls{es} under the context window. A higher $W$ also allows early abnormal samples to warn of later resulting faults, which might otherwise go undetected. Note that we assume the coherence time of observed data is relatively large compared with the context window size.

\subsubsection{Uplink Transmission Phase}\label{sec:frame_phase_2} After aggregating $2W + 1$ samples, the \gls{iot} device enters the uplink data transmission phase. To this end, the \gls{iot} device first compresses accumulated data samples with lossless compression methods, where the resulting data size is denoted as $b_{\mathrm{UL}}^i (W) = b_{H, \mathrm{UL}}^i+b_qN|\mathcal{Q}^i|$ [bits], where $|\mathcal{Q}^i|$ is the cardinality of $\mathcal{Q}^i$, $b_q$ is 
the resolution of sensor measurements, which is set to floating-point 32 bits, and $b_{H, \mathrm{UL}}^i$ is the number of bits for the header and the additional protocol overhead required for uplink transmission at the $i$-th \gls{fd} round. The compressed data is transmitted to the \gls{ap}, e.g., using Wi-Fi. The uplink data transmission time can be defined as $t_{\mathrm{UL}}^i(W) = b_{\mathrm{UL}}^i (W)/R_{\mathrm{UL}}$, where $R_{\mathrm{UL}}$ is the effective uplink data rate.

\subsubsection{Model Training and Link-aware Model Compression}\label{sec:frame_phase_3} When the \gls{es} receives $j$-th sample $\bm{X}_j^i \subseteq \mathcal{Q}^i$ from the \gls{iot} device during the $i$-th \gls{fd} round, it first assigns the label $l_j^i \in \{0, 1\}$, based on whether a fault was reported by the \gls{iot} device owner. Then, the data associated with the correct label is stored in the memory for the $i$-th \gls{fd} round, denoted as $\mathcal{V}_R^i$ as  
$\mathcal{V}_R^i \leftarrow \mathcal{V}_R^i \cup(\bm{X}_j^i, l_j^i)$, where $\mathcal{V}_R^i = \emptyset$ as an initial condition. 
Subsequently, the previous model $\mathcal{M}^i$ is trained with the new training data set $\mathcal{V}_R^i$.
In order to mitigate catastrophic forgetting, we apply the experience replay method considered in \gls{cl} literature~\cite{catastrophic-forgetting}, in which the \gls{es} combines an equal amount of randomly selected older rehearsal data drawn from $\mathcal{R}^i = \bigcup_{i=0}^{i-1}\mathcal{V}_R^i$ with the newly available data $\mathcal{V}_R^i$. 

 For deploying $\mathcal{M}^{i+1}$ to the resource constrained \gls{iot} device, the \gls{es} first compresses the trained \gls{ml} model $\mathcal{M}^{i+1}$. For model compression, the \gls{es} implements successive pruning and quantization operations after training, while taking into account the available link budget $L_Q$. In the pruning operation, the \gls{es} prunes the fraction of $P_L \in [0, 1]$ unimportant weights (the weights whose value is small) in $\mathcal{M}^{i+1}$ by setting its values to zero, while the fraction of $(1-P_L)$ weights remains the same. As for the quantization techniques, we apply post-quantization techniques~\cite{abushahla2025quantized}, in which the \gls{es} quantized all weight and activation values to $Q_L$ [bits]. Finally, the \gls{es} obtains the compressed $i +1$-th \gls{fd} model, $\mathcal{M}^{i+1}$.

\subsubsection{Downlink \gls{ml} Model Transmission Phase}\label{sec:frame_phase_4} After the model has been updated and compressed, $\mathcal{M}^{i+1}$ is further compressed with lossless compression methods resulting in a final model size $b_{\mathrm{DL}}^i(P_L, Q_L)$, which is the function of $P_L$ and $Q_L$. 
Here, the time required for the entire \gls{fd} model reception can be expressed as $t_{\mathrm{DL}}^i(P_L, Q_L) = b_{\mathrm{DL}}^i(P_L, Q_L)/{R_{\mathrm{DL}}}$, where $R_{\mathrm{DL}}$ is the effective downlink data rate. 
After completing the fault detection model update to the \gls{iot} device, it resumes the \gls{fd} phase, using the newly received model $\mathcal{M}^{i+1}$.
\subsection{Energy Model}\label{sec:energy_model}
The \gls{iot} device consumes energy in both the receiving state and transmission state. We denote the power consumption of transmission and reception as $\xi_{\mathrm{tx}}$~[W] and $\xi_{\mathrm{rx}}$~[W], respectively. Here, we ignore the power consumption during the idle-period, such as the power consumed during the \gls{fd} phase for simplifying our analysis. Then, given the transmission and reception time, $t_{\mathrm{UL}}^i(W)$ and $t_{\mathrm{DL}}^i(P_L, Q_L)$, the energy consumed for communication during the $i$-th \gls{fd} round is:
\begin{equation}
 E_{\mathrm{comm}}^i(W, P_L, Q_L) = t_{\mathrm{UL}}^i(W)\xi_{\mathrm{tx}} + t_{\mathrm{DL}}^i(P_L, Q_L)\xi_{\mathrm{rx}}.\label{eq:comm}
\end{equation}

Let us denote the energy required for a single inference task as $E_{\mathrm{inf}}$, which is a sum of energy consumed at the hardware itself, denoted as $E_{\mathrm{HW}}$, and energy consumed by the \gls{dram}, denoted as $E_{\mathrm{DRAM}}$~\cite{moons2017minimum}.\footnote{The specific definition of $E_{\mathrm{HW}}$ and $E_{\mathrm{DRAM}}$ can be found in~\cite{moons2017minimum}.} The number of inference operations conducted by the \gls{iot} device during a single \gls{fd} round, is determined by the transmission condition of the \gls{iot} device, namely by the parameter $\tau_{\mathrm{th}}$. 
Let $I^i(\tau_{\mathrm{th}})$ be the total number of inference operations that the \gls{iot} devices conducted during the $i$-th \gls{fd} phase. Then, the total energy consumed by inference operations during the $i$-th \gls{fd} round can be:
\begin{equation}
 E_{\mathrm{comp}}^i (\tau_{\mathrm{th}}) = E_{\mathrm{inf}}I^i(\tau_{\mathrm{th}}).\label{eq:comp}
\end{equation}

Finally, by using Eqs.~\eqref{eq:comm} and \eqref{eq:comp}, we can define the total energy consumption of the \gls{iot} device at the end of the $i$-th \gls{fd} round:
\begin{equation}
E_{\mathrm{total}}^i(W,\tau_{\mathrm{th}}, P_L, Q_L) =  \sum_{j=1}^i E_{\mathrm{comm}}^j +  E_{\mathrm{comp}}^j.
\end{equation}

\subsection{Inference Accuracy at the \gls{iot} Devices}
We use the recall (true fault discovery rate), defined as a fraction of actual faults, detected at the \gls{iot} devices as described below:
\begin{equation}
\bar{\gamma}_{{R}}^i(P_{L}, Q_{L},W)= \frac{\sum_{j=1}^i \sum_{k=1}^{I^j(\tau_{\mathrm{th}})} 
   \mathbbm{1} (\hat s_k^i = 1)\mathbbm{1} (s_k^i = 1)}{\max(1, A^i)}.\label{eq:average_racall}
\end{equation}
where $A^i$ is the total number of actual fault events, calculated as $A^i= \sum_{j=1}^{i}\sum_{k=1}^{I^j(\tau_{\mathrm{th}})} \mathbbm{1} (s_k^i = 1)$.
\section{Problem Formulation and Solution Approach}
\subsection{Problem Formulation}\label{sec:problem_formulation}
We are now interested in how we can maximize the long term inference accuracy at the \gls{iot} device under the total energy consumption constraint $E_{\mathrm{th}}$ [J]. As the performance of our proposed scheme depends on the transmission policy of the \gls{iot} device, i.e., the transmission threshold $\tau_{\mathrm{th}}$ as well as the compression level of received \gls{ml} model, which is controlled by pruning level $P_L$ and quantization level $Q_L$ in this work, it is desirable to optimize these parameter in terms of average inference accuracy and total energy consumption. The overall problem can be formulated as:
\begin{align}
    \max_{\{P_L, Q_L, W, \tau_{\mathrm{th}}\}}&\quad \bar{\gamma}^i_{{R}}(P_{L}, Q_{L},W)\label{eq:objectiove}\\  
     \text{s.t.}\quad & E_{\mathrm{total}}^i(P_L, Q_L, W, \tau_{\mathrm{th}})\leq E_{\mathrm{th}}, \forall i.\label{eq:energy_constraint} \tag{5a}
\end{align}
To solve this problem, we introduce an approximate solution based on an iterative parameter optimization.

\subsection{Iteration-based Approach}
 Our proposed solution consists of three steps: 1) Optimizing \gls{ml} model compression, i.e., optimizing the set of parameters $\{P_L, Q_L\}$; 2) Optimizing uplink data transmission i.e., the parameter $W$; 3) Optimizing $\tau_{\mathrm{th}}$ for \gls{fd}.

In order to adapt the set of parameters $\{P_L, Q_L\}$ for downlink transmission and the parameter $W$ for uplink transmission under energy constraint $E_\mathrm{th}$ in Eq.~\eqref{eq:energy_constraint}, we introduce the target latency for downlink/uplink transmission, denoted as $\{t_{\mathrm{DL}}^*, t_{\mathrm{UL}}^*\}$. 
To this end, we introduce two reference variables: reference rate $R_{\mathrm{Ref}}$ and reference energy $E_{\mathrm{Ref}}$. The reference energy represents the expected energy consumed at the \gls{iot} device when it applies the \gls{fd} model with $P_L = 0$ and $Q_L = 32$, the maximum size of the context window, $W_{\mathrm{max}}$, and $R_{\mathrm{Ref}}$. 
 Here, we denote the downlink and uplink transmissions time under above mentioned ideal conditions $\tau_\mathrm{DL}={b_{\mathrm{DL}}^i(P_L=0, Q_L=32)}/{R_{\mathrm{Ref}}}$ and $\tau_\mathrm{UL}={b_{\mathrm{UL}}^i (W_{\mathrm{max}})}/{R_{\mathrm{Ref}}}$, respectively. 
Under the energy constraint $E_\mathrm{th}$, the target transmissions times $\{t_{\mathrm{DL}}^*, t_{\mathrm{UL}}^*\}$ are scaled relative to the reference energy $E_{\mathrm{Ref}}$ as follows:
\begin{equation}
    t_{\mathrm{DL}}^*=\tau_{\mathrm{DL}} \left(\frac{E_{\mathrm{th}}-E_{\mathrm{comp}}}{E_{\mathrm{Ref}}}\right),~ t_{\mathrm{UL}}^*=\tau_{\mathrm{UL}}\left(\frac{E_{\mathrm{th}}-E_{\mathrm{comp}}}{E_{\mathrm{Ref}}}\right).
    \label{eq:anchor_eq}
\end{equation}

\subsubsection{Optimizing \gls{ml} Model Compression for Downlink Communication}\label{sec:goodlink}
First, we show how to optimize the compression of the \gls{ml} model, i.e., to find $\{P_L^{*}, Q_L^{*}\}$ that maximizes recall defined in Eq.~\eqref{eq:average_racall}. This is done under the total energy consumption constraint $E_{\mathrm{th}}$ and considering the current link status $L_{Q}$, which is measured at the \gls{es} as the estimated throughput of the link during uplink transmission  $\hat{R}_{\mathrm{UL}}^i = b_{\mathrm{UL}}^i(W)/t_{\mathrm{UL}}^i(W)$. 
Formally, we obtain the optimal pruning with fixed quantization levels to ensure $t^i_{DL}\leq t^*_{DL}$ as:
\begin{equation}
 P_L^{*} = \min~P_L,~\text{s.t., $t_{\mathrm{DL}}^i(P_L, Q_L) \leq t_{\mathrm{DL}}^*$},
\end{equation}
where $t_{\mathrm{DL}}^*$ is target downlink transmission time that depends on the available energy budget $E_{\mathrm{th}}$ as defined in Eq.~\eqref{eq:anchor_eq}.

As analytically characterizing the transmission time for different link conditions is challenging, we model it using linear regression based on the empirical transmission data size $b_{\mathrm{DL}}^i(P_L, Q_L)$ for different pruning ratios $P_{L}^{*} \in [0, 1]$, parametrized by $Q_L$.  

Let downlink package size be $b_{\mathrm{DL}}^i(P_L, Q_L) = a_{\mathrm{DL}}({\mathrm{Q_L}})\,P_L + c_{\mathrm{DL}}({\mathrm{Q_L}})$, where $a_{\mathrm{DL}}({\mathrm{Q_L}})$ is the slope coefficient and $c_{\mathrm{DL}}({\mathrm{Q_L}})$ is a constant value for a given $Q_L$. 
Using the estimated data rate $\hat{R}_{\mathrm{DL}} = \hat{R}_{\mathrm{UL}}^i$, the \gls{es} can decide the optimal pruning rate as follows: 
\begin{equation}
P_{\mathrm{L}}^\star(Q_L) \;=\; \frac{c_{\mathrm{DL}}(\mathrm{Q_L}) - \hat{R}_{\mathrm{DL}}\cdot t_{\mathrm{DL}}^*}{a_{\mathrm{DL}}(\mathrm{Q_L})},~Q_L \in \{8, 32\},\label{eq:pruning}
\end{equation}
where we substitute $b_{\mathrm{DL}}^i(P_L, Q_L) = t_{\mathrm{DL}}^i(P_L, Q_L)\hat{R}_{\mathrm{DL}}$ and $t_{\mathrm{DL}}^i(P_L, Q_L) = t_{\mathrm{DL}}^*$.

As demonstrated in~\cite{pruningvsquan}, increasing $P_{\mathrm{L}}$ degrades model accuracy more than achieving an equivalent model size through quantization by lowering $Q_{\mathrm{L}}$.  Considering this, we propose a heuristic-based optimal parameter $\{P_\mathrm{L}^\star, Q_\mathrm{L}^\star\}$ selection method  in Eq.~\eqref{eq:pruning} that achieves high inference accuracy while satisfying the target transmission time, as defined in Eq.~\eqref{eq:anchor_eq}. We select the optimal parameters based on the predetermined pruning threshold $P_{\mathrm{th}}$. Here, the value of $P_{\mathrm{th}}$ is selected such that it satisfies $b_{\mathrm{DL}}(P_{\mathrm{th}},Q_L = 32) = b_{\mathrm{DL}}(P_L=0,Q_L=8)$.
Then, if $P_{\mathrm{L}}^\star(Q_L = 32) \leq P_{\mathrm{th}}$, the \gls{es} applies $\{P_\mathrm{L}^\star, Q_\mathrm{L}^\star\} = \{P_\mathrm{L}^*(Q_L = 32), 32\}$. On the other hand, if $P_{\mathrm{L}}^\star(Q_L = 32) > P_{\mathrm{th}}$, the \gls{es} applies $\{P_\mathrm{L}^\star, Q_\mathrm{L}^\star\}= \{P_\mathrm{L}^*(Q_L = 8), 8\}$.

\subsubsection{Optimizing Uplink Data Transmission}\label{sec:opt_for_W}
Similarly to the approach mentioned in Sec.~\ref{sec:goodlink}, we adapt linear regression to obtain the optimal context window size $W^{*}$. Specifically, we first obtain $b_{\mathrm{UL}}^i (W)$ for $W$ empirically. Then, to obtain the relationship between $b_{\mathrm{UL}}^i (W)$ and $W$, we apply linear regression $  b_{\mathrm{UL}}^i (W) = a_{\mathrm{UL}}\,W + c_{\mathrm{UL}}$, where $a_{\mathrm{UL}}$ is the slope coefficient and $c_{\mathrm{UL}}$ is a constant value. 

Given the estimated uplink data rate $\hat{R}_{\mathrm{UL}}$, the \gls{iot} device can adjust the context window size as follows:
\begin{equation}\label{eq:rate_to_n}
W^\star \;=\; \frac{c_{\mathrm{UL}} - \hat{R}_{\mathrm{UL}}\cdot t^*_{\mathrm{UL}}}{a_{\mathrm{UL}}},
\end{equation}
where $b_{\mathrm{UL}}^i(W) = t_{\mathrm{UL}}^i(W)\hat{R}_{\mathrm{UL}}$ and $t_{\mathrm{UL}}^i(W) = t_{\mathrm{UL}}^*$.

\subsubsection{Optimizing Decision Threshold $\tau_{\mathrm{th}}$}\label{sec:fault_detectors} 
We obtain the optimal fault decision threshold $\tau_{\mathrm{th}}^{*}$ for each pair of parameters $\{P_{L}^*, Q_{L}^*,W^*\}$ calculated in Secs.~\ref{sec:goodlink} and ~\ref{sec:opt_for_W}, by using a full \gls{cl} operation based on the system model described in Sec.~\ref{sec:sys_model}.
For each $\tau_{\mathrm{th}} \in [0,1], \Delta\tau_{\mathrm{th}}=0.1$, 
we record the \gls{tpr} and \gls{fpr} on the fault detection task, which is denoted as $\mathrm{TPR}(\tau)$, $\mathrm{FPR}(\tau)$ and select the optimal $\tau_{\mathrm{th}}^{*}$ based on a \gls{roc} curve as follows: 
\begin{equation}\label{eq:optimumtau}
    \tau_{\mathrm{th}}^{*} = 
\arg\min_{\tau \in (0, 1)}
\left( 
    \mathrm{FPR}(\tau)^{2} 
    + 
    \bigl(1 - \mathrm{TPR}(\tau)\bigr)^{2}
\right).
\end{equation}

\section{Experimental Setup and Results}
\subsection{Experimental Setup}
We consider a setup with a transmitter (\gls{iot} device) and receiver (\gls{es}) pair, using two Ubuntu~24.04 computers. These two nodes communicate via a Wi-Fi router with a wireline connection to \gls{es} and a wireless to \gls{iot}.
The \gls{iot} device and the \gls{es} operation is based on the description presented in Sec.~\ref{sec:sys_model}. To ensure reliable data/model transmission, we use the \gls{tcp} on the transport layer.

We exploit a practical fault data set pump\_sensor\_data~\cite{used-dataset}. It has about 220k samples, each of which includes 50 features corresponding to typical pump telemetry data, along with the correct label from the set \{NORMAL,\,~BROKEN,\,~RECOVERING\}. Here, the ``NORMAL" and ``RECOVERING" labels correspond to the non-faulty state, while the ``BROKEN" label corresponds to the fault state.
The data set includes only 7 BROKEN labels, i.e., fault events are rare.  At the $0$-th communication round, we split the data set into two sets:  $10 \%$ of the data for training and the remaining $90 \%$ for testing.
We ensure that all faults remain in the test set. 

We consider \gls{ae} and \gls{mlp} as fault detectors. The \gls{mlp} uses a sigmoid output, and the \gls{ae} a linear output. 
 The loss function for training the \gls{ae} is:
\begin{equation}\label{eq:autoencloss}
    \mathcal{L}({\bf{x}_k^i},{\hat{\bf{x}}_k^i})=\left[\lambda_{1}\,\mathbbm{1}(s_k^i = 1) +\lambda_{0}\mathbbm{1}(s_k^i = 0)\right] \frac{||{\bf{x}_k^i}-{\hat{\bf{x}}_k^i}||_2^2}{N},
\end{equation}
where $\bf{x}_k^i,\hat{\bf{x}}_k^i$ correspond to the given sample and its reconstruction from the \gls{ae}, $\lambda_{1}$ and $\lambda_{0}$ represent weights for the faulty and normal labeled data, respectively, which is set to $\lambda_{1}= -0.1$ and $\lambda_{0}= 1$. The setting of $\lambda_{1}$ to be negative pushes the weights in the opposite direction when training, preventing the model from learning to reconstruct those samples~\cite{negativelambda}.
Further, we use the TensorFlow Lite framework~\cite{tflite-paper} for each model deployment. We set the number of training epochs $L_{\mathrm{epoch}}$ for each \gls{fd} round as $L_{\mathrm{epoch}}=2000/(2W+1)$, resulting in $L_{\mathrm{epoch}} \in [5,16]$.

The power consumption in the receiving/transmitting state is set to, respectively, $\xi_{\mathrm{rx}}= 0.33$~[W] and $\xi_{\mathrm{tx}} = 0.79$~[W] based on the specifications of the ESP32~\cite{ESP-32-pow-Con-article}. 
The parameters for the \gls{ae} model are: total weights and biases $N_s = 34298$, total multiply and accumulate operations $N_c = 33792$, and total activations $A_s = 506$. Based on this $E_{\mathrm{inf}}$ can be calculated (as described in Sec.~\ref{sec:energy_model}) as $E_{\mathrm{inf}}=1.4$ $\mu$J for $Q_L = 8$ and $E_{\mathrm{inf}}=6.6~\mu$J for $Q_L = 32$. 
Setting reference rate $R_\mathrm{Ref}=1$ Mbps results in $E_\mathrm{Ref}=60J$.
Finally, $t_{\text{UL}}$/$t_{\text{DL}}$ are measured on each device, taking into account the transmission data size as well as protocol overhead to obtain a realistic transmission/reception time.

We introduce two comparison schemes: 1) State of the art non-adaptive \gls{cl} scheme ``Hawk"~\cite{hawk}  and 2) Periodic Sampling. In Hawk, we apply constant values for the model compression and data transmission scheme, i.e., $P_L = 0, Q_L = 32$, and $W=200$, without considering the current link status and the available energy budget. In this scheme, if the energy constraint is not satisfied, i.e., if $ E_{\mathrm{total}} > E_{\mathrm{th}}$, the system no longer updates the \gls{ml} model and stops the \gls{fd} task. 
On the other hand, in Periodic Sampling, the \gls{iot} device periodically picks samples to designate as faults irrespective of their importance for the model improvement with context window size $W = 200$. Here, we use the optimal periodicity for sampling based on $E_{\mathrm{th}}$, which satisfies $E_{\mathrm{total}}= E_{\mathrm{th}}$.

\subsection{Comparison of Classifier ROC Curves}
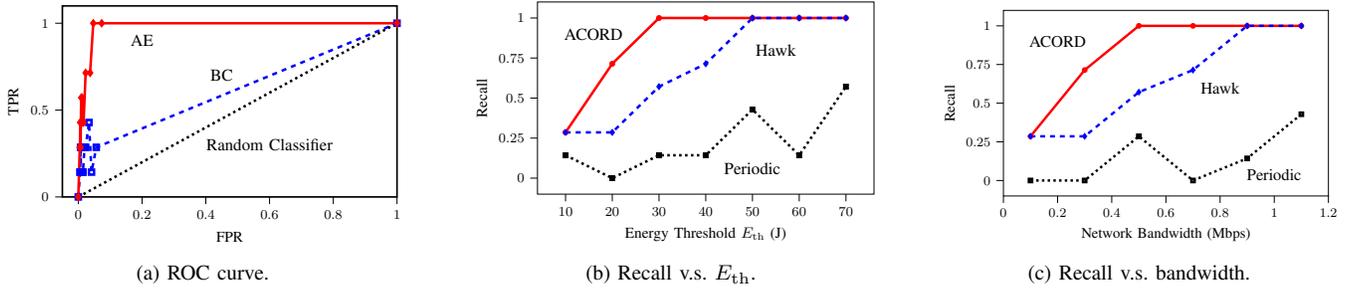
\begin{figure*}[!t]
    \centering

    \subfloat[\Gls{roc} curve.\label{fig:model-comparison_ROC}]{
        \resizebox{0.3\textwidth}{!}{\input{ROC_curves_improved}}
    } \hfill
    \subfloat[Recall v.s. $E_{\mathrm{th}}$.\label{fig:fulladaptation_ene_change}]{
        \resizebox{0.3\textwidth}{!}{\input{algo-comp-bandwidth}}
    }\hfill
    \subfloat[Recall v.s. bandwidth.\label{fig:fulladaptation_bandwidth}]{
        \resizebox{0.3\textwidth}{!}{\input{bandwidth_comparison}}
    }\hfill

    \caption{(a) Comparison of \gls{roc} curves for \gls{ae}, \gls{mlp} and random classifier with $P_L = 0, Q_L = 0,W = 200$; Recall of ACORD with Hawk and Periodic Sampling 
    (b) as a function of energy threshold $E_{\mathrm{th}}$ and (c) as a function of network bandwidth.}
    \label{fig:acord_subfigures_wide}
\end{figure*}

Fig.~\ref{fig:model-comparison_ROC} shows the achievable set of \gls{tpr} and \gls{fpr}, for \gls{ae}, \gls{mlp}, and a random classifier. The random classifier declares the fault with predetermined probability, failing to control the balance of \gls{tpr} and \gls{fpr}, as it declares fault without considering the relevance of observations for the fault event. 
Comparing the performance of the \gls{ae} and the \gls{mlp}, we can clearly see that the \gls{ae} achieves significantly higher \gls{tpr} than the \gls{mlp} approach, while maintaining small \gls{fpr}. This is because the \gls{ae} can learn from both \gls{fp} and \gls{tp} data during the training process, i.e., it can learn to reconstruct \glspl{fp} (\glspl{tp}) better (worse) due to the use of the dual function loss from Eq.~\eqref{eq:autoencloss}. On the other hand, the \gls{mlp} cannot learn from \glspl{fp} until some \glspl{tp} have been collected, so as not to overfit to the \glspl{fp}. This shows the advantage of the \gls{ae} for the fault detection task; thus, we use \gls{ae}, hereafter.

\subsection{Comparison of ACORD with the Baseline Schemes}
Fig.~\ref{fig:fulladaptation_ene_change} shows recall for optimized \gls{acord}, Hawk and Periodic Sampling as a function of energy constraint $E_{\mathrm{th}}$, where we set the network bandwidth to 1~Mbps. 
We can observe that the recall becomes larger as the energy constraint $E_{\mathrm{th}}$ increases. The higher $E_{\mathrm{th}}$ allows more \gls{cl} rounds with an increased number of \glspl{fp}. This increases the available training data set at the \gls{es}, which increases the recall due to the improvement of model accuracy during the model training phase. 
Fig.~\ref{fig:fulladaptation_ene_change} also shows that \gls{acord} achieves higher recall than the baseline schemes, especially when energy constraint $E_{\mathrm{th}}$ is small. With Periodic Sampling, the energy constraint is always satisfied by adjusting the rate of data transmission, but the \gls{iot} device transmits data irrespective of its importance, deteriorating the recall. Hawk~\cite{hawk} achieves informative data updates by exploiting the installed model, but this scheme does not consider the available energy budget and could spend most of the energy for data sharing and model updates in the early rounds, exceeding the energy budget relatively early. 
On the other hand, \gls{acord} realizes informative data/model sharing, taking into account the available energy budget by adjusting the transmission data size and \gls{ml} model compression size. Thanks to the energy-frugal informative data/model transmission, \gls{acord} can keep using the installed model for the \gls{fd} task longer than Hawk, which leads to the higher recall.

Fig.~\ref{fig:fulladaptation_bandwidth} shows recall for optimal \gls{acord}, Hawk and Periodic Sampling as a function of network bandwidth, where we set $E_{\mathrm{th}} = 60$~J. From this figure, we can first see that for all schemes the recall increases as network bandwidth becomes larger. This is because higher bandwidth allows for more transmissions within the same energy budget.  
Next, we can see that \gls{acord} achieves higher recall than Hawk and Periodic Sampling, especially when the network bandwidth is small. 
This is because \gls{acord} can control both transmission and reception time to adjust to the link status using Eq.~\eqref{eq:anchor_eq}, while Hawk and Periodical Sampling use energy inefficiently as they do not consider the available energy and/or link status.

These results clearly demonstrate the importance of designing model/data transmission schemes which take the link and energy budget at the \gls{iot} device into account.

\section{Conclusion}
This paper introduced ACORD, an event-driven \gls{cl} and communication framework for resource-constrained \gls{iot} networks focusing on a \gls{fd} task at the \gls{iot} device. The proposed framework was designed to continuously improve a \gls{fd} model through interactions between the \gls{iot} device and the \gls{es} via communication. Furthermore, we have proposed a link-aware model compression and \gls{iot} data transmission method, in which the \gls{es} and \gls{iot} devices tune the model/data size based on the estimated link status. 
The experiments confirmed that the proposed approach can provide high inference accuracy while satisfying the energy constraint compared to the baseline schemes, especially when the available energy budget and network bandwidth are limited.

Our future work includes the design of \gls{ml} model transmission and \gls{iot} data transmission policies for both the \gls{es} and \gls{iot} devices, in which multiple nodes contend for the channel when transmitting data. 

\bibliographystyle{IEEEtran}
\bibliography{IEEEabrv,refs}
\end{document}

%% file: acronyms.tex
\newacronym{acord}{ACORD}{Adaptive Compression Online Resource-aware fault Detection}
\newacronym{adc}{ADC}{analog-to-digital converter}
\newacronym{dram}{DRAM}{dynamic random access memory}
\newacronym{dnn}{DNN}{deep neural network}
\newacronym{iot}{IoT}{internet of things}
\newacronym{mcu}{MCU}{micro controller unit}
\newacronym{ml}{ML}{machine learning}
\newacronym{cl}{CL}{continual learning}
\newacronym{lq}{$L_Q$}{network link status}
\newacronym{ae}{AE}{autoencoder}
\newacronym{ap}{AP}{access point}
\newacronym{auc}{AuC}{area under the curve}
\newacronym{wur}{WuR}{wake-up radio}
\newacronym{fd}{FD}{fault detection}
\newacronym{fl}{FL}{Federated Learning}
\newacronym{mlp}{BC}{binary classifier}
\newacronym{tp}{TP}{true positive}
\newacronym{tpr}{TPR}{true positive rate}
\newacronym{fp}{FP}{false positive}
\newacronym{fpr}{FPR}{false positive rate}
\newacronym{es}{ES}{edge server}
\newacronym{tcp}{TCP}{Transmission Control Protocol}
\newacronym{tdr}{TDR}{true (false) discovery rate}
\newacronym{wsn}{WSN}{wireless sensor network}
\newacronym{pid}{PID}{proportional–integral–derivative}
\newacronym{kpi}{KPI}{key performance indicator}
\newacronym{roc}{ROC}{receiver operating characteristic curve}

%% file: ROC_curves_improved.tex
\begin{tikzpicture}
\definecolor{darkgray176}{RGB}{176,176,176}
\definecolor{lightgray204}{RGB}{204,204,204}
\definecolor{navy}{RGB}{0,0,128}

\begin{axis}[
    width=7cm,
    height=4cm,
    xlabel={FPR},
    ylabel={TPR},
    xmin=-0.05, xmax=1,
    ymin=0, ymax=1.1,
    tick label style={},
    label style={},
    line width=1pt,
    mark size=1.5pt,
    legend style={
        at={(0.5,-0.45)},
        anchor=north,
        draw=none,
        fill=none,
        legend columns=1
    }
]

\addplot [color=blue, line width=1.5pt, dashed, mark=square, mark size=1.5, mark options={solid}]
table[col sep=space] {ROC_BC.dat};
\node[black] at (axis cs:0.45,0.7) {BC};

\addplot [color=red, line width=1.5pt, mark=diamond, mark size=1.5, mark options={solid}]
table[col sep=space] {ROC_AE.dat};
\node[black] at (axis cs:0.2,0.9) {AE};

\addplot [line width=1.5pt, black, dotted]
table[col sep=space] {ROC_RC.dat};
\node[black] at (axis cs:0.6,0.3) {Random Classifier};

\end{axis}
\end{tikzpicture}

%% file: algo-comp-bandwidth.tex
\begin{tikzpicture}

\definecolor{crimson2143940}{RGB}{214,39,40}
\definecolor{steelblue31119180}{RGB}{31,119,180}

\begin{axis}[
    width=7cm,
    height=4cm,
    xlabel={Energy Threshold $E_{\mathrm{th}}$ (J)},
    ylabel={Recall},
    ytick={0, 0.25, 0.5, 0.75, 1},
    scaled ticks=false,
    grid style={line width=.2pt, draw=gray!30},
    tick label style={},
    label style={},
    tick align=outside,
    legend style={
        at={(0.5,-0.5)},
        anchor=north,
        draw=none,
        fill=none,
        legend columns=2
    }
]

\addplot[
    solid,
    red,
    mark=o,
    line width=1.5pt,
    mark size=1,
    mark options={solid},
    table/y expr=\thisrowno{2}/7
]
table[col sep=space]{E_acord.dat};
\node[black] at (axis cs:16,0.9) {ACORD};

\addplot[
    blue,
    dashed,
    mark=diamond,
    line width=1.5pt,
    mark size=1,
    mark options={solid},
    table/y expr=\thisrowno{2}/7
]
table[col sep=space]{E_hawk.dat};
\node[black] at (axis cs:55,0.8) {Hawk};

\addplot[
    black,
    dotted,
    mark=square,
    line width=1.5pt,
    mark size=1,
    mark options={solid},
    table/y expr=\thisrowno{2}/7
]
table[col sep=space]{E_periodic.dat};
\node[black] at (axis cs:50,0.06) {Periodic};

\end{axis}
\end{tikzpicture}

%% file: bandwidth_comparison.tex
\begin{tikzpicture}

\definecolor{crimson2143940}{RGB}{214,39,40}
\definecolor{steelblue31119180}{RGB}{31,119,180}

\begin{axis}[
    width=7cm,
    height=4cm,
    xlabel={Network Bandwidth (Mbps)},
    ylabel={Recall},
    ytick={0, 0.25, 0.5, 0.75, 1},
    scaled ticks=false,
    grid style={line width=.2pt, draw=gray!30},
    tick label style={},
    label style={},
    tick align=outside,
    legend style={
        at={(0.5,-0.5)},
        anchor=north,
        draw=none,
        fill=none,
        legend columns=2
    }
]

\addplot[
    solid, red, mark=o,
    line width=1.5pt, mark size=1,
    mark options={solid},
    table/y expr=\thisrowno{2}/7
]
table[col sep=space]{BW_acord.dat};
\node[black] at (axis cs:0.2,0.9) {ACORD};

\addplot[
    blue, dashed, mark=diamond,
    line width=1.5pt, mark size=1,
    mark options={solid},
    table/y expr=\thisrowno{2}/7
]
table[col sep=space]{BW_hawk.dat};
\node[black] at (axis cs:0.8,0.6) {Hawk};

\addplot[
    black, dotted, mark=square,
    line width=1.5pt, mark size=1,
    mark options={solid},
    table/y expr=\thisrowno{2}/7
]
table[col sep=space]{BW_periodic.dat};
\node[black] at (axis cs:1,0.038) {Periodic};

\end{axis}
\end{tikzpicture}